\documentclass[runningheads]{llncs}
\usepackage[T1]{fontenc}
\usepackage{graphicx,verbatim}
\usepackage{hyperref}
\usepackage{color}

\usepackage{xcolor}
\usepackage{adjustbox}
\usepackage{booktabs}
\usepackage{amsmath}
\usepackage{amssymb}
\usepackage{pifont}
\usepackage{multicol,multirow}
\usepackage{marvosym}
\newcommand{\cmark}{\ding{51}}%
\newcommand{\xmark}{\ding{55}}%
\usepackage[capitalize]{cleveref}
\crefname{table}{Tab.}{Tabs.}

\begin{document}
\title{Adapting Vision Foundation Models for Real-time Ultrasound Image Segmentation}
%
\author{Xiaoran Zhang\thanks{\scriptsize This work was carried out during the internship of X. Zhang at United Imaging Intelligence.}\inst{1} \and
Eric Z. Chen\inst{2} \and
Lin Zhao\inst{2} \and
Xiao Chen\inst{2} \and
Yikang Liu\inst{2} \and
Boris Maihe\inst{2} \and
James S. Duncan\inst{1,3} \and
Terrence Chen\inst{2} \and
Shanhui Sun\inst{2}$^{(\text{\Letter})}$
} 
\authorrunning{X. Zhang et al.}
%
\institute{Biomedical Engineering, Yale University, New Haven, USA \\
\email{xiaoran.zhang@yale.edu}\\
\and
United Imaging Intelligence, Burlington, USA\\
\email{shanhui.sun@uii-ai.com}\\
\and
Radiology \& Biomedical Imaging, Yale University, New Haven, USA}


    
\maketitle              

\begin{abstract}
We propose a novel approach that adapts hierarchical vision foundation models for real-time ultrasound image segmentation. Existing ultrasound segmentation methods often struggle with adaptability to new tasks, relying on costly manual annotations, while real-time approaches generally fail to match state-of-the-art performance. To overcome these limitations, we introduce an adaptive framework that leverages the vision foundation model Hiera to extract multi-scale features, interleaved with DINOv2 representations to enhance visual expressiveness. These enriched features are then decoded to produce precise and robust segmentation. We conduct extensive evaluations on six public datasets and one in-house dataset, covering both cardiac and thyroid ultrasound segmentation. Experiments show that our approach outperforms state-of-the-art methods across multiple datasets and excels with limited supervision, surpassing nnUNet by over 20\% on average in the 1\% and 10\% data settings. Our method achieves $\sim$77 FPS inference speed with TensorRT on a single GPU, enabling real-time clinical applications.


\end{abstract}

\section{Introduction}
\label{sec:intro}




Ultrasound image segmentation is a long-standing challenge in medical image analysis, playing a crucial role in applications such as robot-assisted imaging \cite{jiang_robotic_2023,salcudean_robot-assisted_2022}, cardiac function analysis \cite{ta_multi-task_2024,leonardis_adaptive_2025,linguraru_heteroscedastic_2024}, and real-time disease monitoring \cite{loizou_review_2014,amiri_fine-tuning_2020}. However, ultrasound images are difficult to segment due to their low signal-to-noise ratio, speckle noise, and high anatomical variability across patients \cite{michailovich_despeckling_2006}. Furthermore, anatomical boundaries in ultrasound images are often indistinct and ambiguous, complicating precise delineation. Even for experienced annotators, inter-observer variability in ultrasound image segmentation is notably higher than in other modalities such as MRI or CT \cite{bernard_deep_2018,webb_comparing_2021,zhou_cardiac_2017}, posing challenges for automatic segmentation models to achieve accuracy and consistency.c

In response to these challenges, deep learning-based approaches have made significant progress in ultrasound image segmentation, with CNN-based \cite{ronneberger_u-net_2015,isensee_nnu-net_2021} and transformer-based architectures \cite{greenspan_mednext_2023,greenspan_swinunetr-v2_2023} trained in a supervised manner. However, these models often struggle to generalize to unseen ultrasound distributions, making deployment in diverse clinical scenarios challenging. Their adaptability to new classes and tasks is further constrained by the reliance on costly, task-specific annotations. Moreover, while some real-time approaches \cite{ou_rtseg-net_2024,vaze_low-memory_2020} improve efficiency, they typically fail to achieve state-of-the-art performance.


Recently, vision foundation models \cite{awais_foundation_2025,tong_cambrian-1_2024} have emerged as a promising alternative to overcome these challenges. Pre-trained on large-scale, general-purpose datasets, these models capture broad visual representations, enabling them to generalize across diverse tasks and imaging domains \cite{dosovitskiy_image_2020}. 
Inspired by segmentation-specialized foundation models such as SAM \cite{kirillov_segment_2023}, several approaches \cite{deng_memsam_2024,ravishankar_sonosam_2023,lin_beyond_2024} have been proposed to adapt SAM for ultrasound segmentation. However, these methods are often constrained by single-scale feature extraction, despite the crucial role of multi-scale representation in segmentation \cite{ronneberger_u-net_2015,chen_transunet_2021}. Furthermore, they do not exploit complementary vision foundation models like DINOv2 \cite{oquab_dinov2_2023}, which capture rich semantic representations crucial for ultrasound image segmentation, where anatomical boundaries are often indistinct and ambiguous. 

To address these limitations, we propose a novel approach that adapts hierarchical encoder of vision foundation model Hiera \cite{ryali_hiera_2023}, which not only generates multi-scale features but also maximizes efficiency, achieving up to 2.3$\times$ speedup over traditional ViTs \cite{ryali_hiera_2023}. These multi-scale features are further interleaved with DINOv2 features to leverage visual semantics and enhance visual expressiveness. The enriched features are then decoded to generate precise and robust segmentations. Our method is evaluated on six public datasets and one in-house dataset, spanning cardiac and thyroid ultrasound segmentation. We demonstrate that our approach outperforms existing methods on public benchmarks, achieving state-of-the-art performance while maintaining real-time inference speeds.

\textbf{Our contributions} are as follows:  
(1) We propose a novel approach that adapts Hiera encoders and integrates DINOv2 features through feature interleaving to enhance visual representation for improved ultrasound segmentation.    
(2) Our method shows strong generalization under limited supervision, outperforming nnUNet by over 20\% on average in cardiac segmentation when all trained on 1\% and 10\% of the data.
(3) We achieve state-of-the-art performance on CAMUS and TN3K based on region-overlap metrics while consistently outperforming baselines across other datasets. Additionally, our method enables real-time inference at $\sim$77 frames per second (FPS) with TensorRT on a single GPU.  

\section{Related works}
\label{sec:related}

\noindent\textbf{Ultrasound image segmentation} has been extensively studied and can be broadly categorized into CNN-based approaches \cite{ronneberger_u-net_2015,isensee_nnu-net_2021,ansari_dense-psp-unet_2023,stock_segment_2024,ta_multi-task_2024} and Transformer-based approaches \cite{chen_transunet_2021,wu_cross-image_2023,ansari_advancements_2024}. More recently, vision foundation models like SAM \cite{kirillov_segment_2023} have been explored for medical image segmentation due to their strong generalization capabilities \cite{ma_segment_2024}. For instance, SonoSAM \cite{ravishankar_sonosam_2023} and SAMUS \cite{lin_beyond_2024} adapt SAM for ultrasound segmentation, while MemSAM \cite{deng_memsam_2024} extends its application to ultrasound videos by incorporating a memory attention mechanism. However, these SAM-based approaches primarily operate on a single feature scale, limiting their ability to capture multi-scale contextual information. In contrast, we adapt Hiera \cite{ryali_hiera_2023} in SAM2 \cite{ravi_sam_2024} for multi-scale feature extraction and integrate DINOv2 \cite{oquab_dinov2_2023} image encoder to enhance visual feature representation, providing a richer and more robust understanding of ultrasound images.


\noindent\textbf{Real-time ultrasound segmentation} in clinical applications predominantly relies on CNN-based architectures \cite{looney_fully_2018,vaze_low-memory_2020,dorent_patient-specific_2024}, with most efforts focused on optimizing UNet by making it more lightweight \cite{howell_deep_2024} and improving robustness to distribution shifts \cite{ou_rtseg-net_2024}. However, these methods often fall short of achieving state-of-the-art performance and lack integration with recent advancements in vision foundation models and segmentation techniques. In contrast, our method not only attains state-of-the-art performance but also maintains computational efficiency, achieving $\sim$77 FPS during inference when converted to TensorRT, making it well-suited for real-time clinical applications.

\noindent\textbf{Parameter-efficient fine-tuning of vision foundation models} optimizes adaptability while minimizing computational cost. Inspired by prefix-tuning in NLP \cite{lester_power_2021}, visual prompt tuning \cite{avidan_visual_2022} prepends learnable embeddings to image patches. LoRA \cite{hu_lora_2021} and its variants \cite{qiu_controlling_2023,yeh_navigating_2024} modify low-rank attention weights for efficient adaptation. Adapters, small trainable modules in frozen networks, enhance flexibility \cite{rebuffi_learning_2017}, with recent work improving the parameter-accuracy trade-off \cite{steitz_adapters_2024}. We extend these ideas by designing adapters for hierarchical encoder and developing a multiscale decoder for optimal performance-efficiency balance.


\section{Methods}
\label{sec:methods}
\begin{figure}[tb]
    \centering
    \includegraphics[scale=0.05]{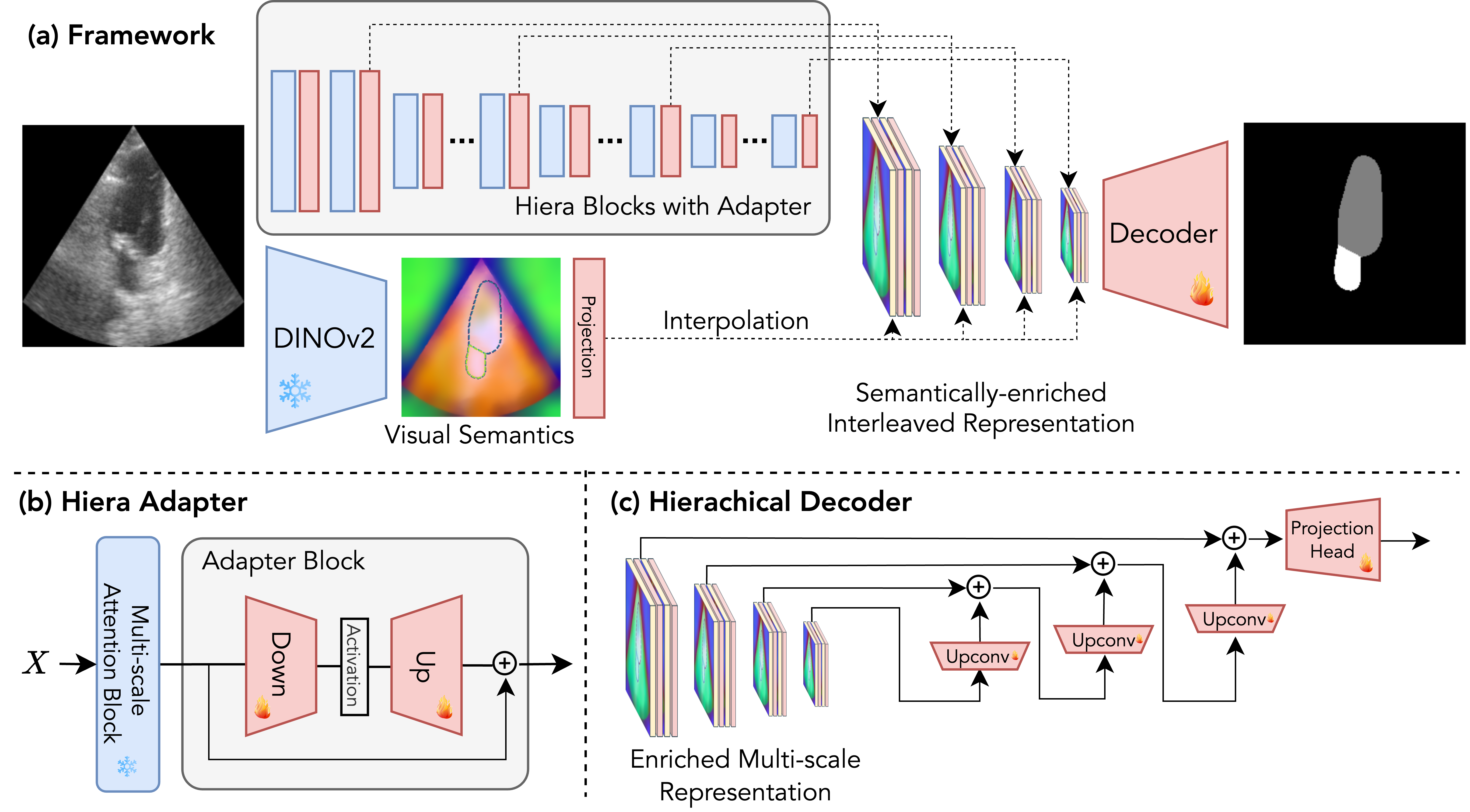}
    \caption{We adapt Hiera to extract multi-scale features, interleaved with DINOv2 features and decoded by a hierarchical decoder. Red blocks denote trainable parameters.} 
    \label{fig:main_framework}
\end{figure}

Let $I\in\mathbb{R}^{H\times W}$ be a 2D ultrasound image with imaging function $I: \Omega\to [0,1]$ over domain $\Omega$. Our goal is to predict a pixel-wise segmentation map $\bar{S}\in\{0, \dots, C-1\}^{H \times W}$ using our framework (\cref{fig:main_framework} (a)). The model outputs logits $\hat{S}\in\mathbb{R}^{C\times H \times W}$, where the final segmentation is obtained via $\arg\max$ over the class dimension, aiming for $\bar{S}$ to match ground truth $S$. 
As shown in \cref{fig:main_framework} (a), the framework comprises (1) a Hiera adapter, (2) interleaved DINOv2 features in the encoding pathway, and (3) a hierarchical decoder that processes multi-scale features to produce pixel-wise logits.

\noindent\textbf{Hiera adapter.}
We introduce a lightweight adapter (\cref{fig:main_framework} (b)) positioned after the skip connection in Hiera’s multi-scale attention block built on MViTv2 \cite{li_mvitv2_2022}. Let $X \in \mathbb{R}^{b \times h \times w \times d}$ be the output of this block, where $b$ is the batch size, $(h, w)$ are spatial dimensions, and $d$ is hidden dimension. The adapter follows a bottleneck structure with a skip connection and is defined as:
\begin{equation}
    \text{Adapter}(X) = \sigma\left(XW_{\text{down}} + b_{\text{down}}\right)W_{\text{up}} + b_{\text{up}} + X,
\end{equation}
where $W_{\text{down}} \in \mathbb{R}^{d \times r}$ and $b_{\text{down}} \in \mathbb{R}^{r}$ are the weights and biases of the down-projection linear layer, while $W_{\text{up}} \in \mathbb{R}^{r \times d}$ and $b_{\text{up}} \in \mathbb{R}^{d}$ correspond to the up-projection layer. $\sigma(\cdot)$ denotes activation function. We set $r = d/4$ and use GELU activation. During training, only the adapter parameters are updated, enabling efficient fine-tuning. The Hiera encoder extracts multi-scale features at $N$ levels $f_{\text{hiera}}(I) = \{F_1^h, F_2^h, \dots, F_N^h\}$,
where $F_n^h \in \mathbb{R}^{b \times h_n \times w_n \times d_{\text{hiera}}}$ represents feature at scale $n$, with spatial dimensions $(h_n, w_n)$ and feature dimension $d_{\text{hiera}}$. 

\noindent\textbf{Feature interleaving.}
To enhance semantic representation, we incorporate an auxiliary DINOv2 encoder. 
We do not finetune or adapt DINOv2, as its large-scale self-supervised contrastive training equips it with a strong ability to capture generalized structure and texture, ensuring robust generalization across diverse image distributions.
The extracted features $f_{\text{dino}}(I) \in \mathbb{R}^{b \times h \times w \times d_{\text{dino}}}$ are resized via bilinear interpolation to match the spatial dimensions at each scale $n$. Instead of concatenation, we apply an interleaving strategy by merging features slice by slice along channel dimensions $F_n = \text{Interleave}(F_n^h, F_n^d)$,
where $F_n^d \in \mathbb{R}^{b \times h_n \times w_n \times d_{\text{dino}}}$ are the spatially aligned DINOv2 features. Since the number of channels in DINOv2 and Hiera do not match ($d_{\text{dino}} \neq d_{\text{hiera}}$), we introduce a projection layer with weights $W_{\text{proj}} \in \mathbb{R}^{d_{\text{dino}} \times d_{\text{hiera}}}$ and biases $b_{\text{proj}} \in \mathbb{R}^{d_{\text{hiera}}}$ to align feature channels before interleaving. Final hierarchical representation is:
\begin{equation}
    F = \{F_1, \dots, F_N\}, \quad F_n \in \mathbb{R}^{b \times h_n \times w_n \times 2d_{\text{hiera}}} \label{eq:feats}. 
\end{equation}
By feature interleaving, we ensure a more fine-grained fusion of local (Hiera) and global (DINOv2) features, improving feature expressiveness across scales.

\noindent\textbf{Hierarchical decoder.}
To decode the multi-scale features from \cref{eq:feats}, we propose a hierarchical decoder that progressively fuses coarse-to-fine representations. At each scale, it upsamples the coarse feature map via transposed convolution, integrates finer-scale features through concatenation, and refines them with a convolutional block. This ensures effective spatial propagation while preserving multi-scale context (\cref{fig:main_framework} (c)). The final projection head combines upsampling and convolution layers to refine features to the original resolution, reconstructing dense pixel-wise logits while balancing accuracy and computational efficiency.

\section{Experiments}
\label{sec:experiments}


\noindent\textbf{Datasets and preprocessing.} 
We evaluate our approach on six public and one in-house 2D ultrasound datasets covering cardiac and thyroid segmentation. For \textbf{cardiac}, we extracted end-diastole and end-systole frames in two- and four-chamber views with left ventricle and atrium annotations: (1) CAMUS \cite{leclerc_deep_2019} (train/val/test = 1,200/400/400 cases from 500 patients), (2) CardiacUDA \cite{yang_graphecho_2023} (484/105/145 from 149 patients in Site G). For \textbf{thyroid}, we followed preprocessing in \cite{gong_multi-task_2021} and evaluated (3) TN3K \cite{gong_multi-task_2021} (2,303/576/614), (4) DDTI \cite{pedraza_open_2015} (637 test cases for TN3K-trained models), (5) Stanford \cite{noauthor_thyroid_nodate} (10,358/3,453/3,453), (6) TG3K \cite{gong_multi-task_2021} (2,580/646/359), and (7) an in-house dataset (17,388/2,700/3,280). All images were resized to $224 \times 224$.

\noindent\textbf{Baselines.} 
We evaluated CNN-based methods: (1) UNet \cite{ronneberger_u-net_2015}, (2) nnUNet \cite{isensee_nnu-net_2021}; transformer-based: (3) MedNeXt \cite{greenspan_mednext_2023}, (4) SwinUNETR \cite{greenspan_swinunetr-v2_2023}; and SAM-based approaches: (5) SAMUS \cite{lin_beyond_2024}, (6) MedSAM2 \cite{zhu_medical_2024}. We also included MemSAM \cite{deng_memsam_2024} for CAMUS and SHAN \cite{linguraru_shan_2024} for TN3K. All baselines used default settings.


\noindent\textbf{Implementation details.} 
All experiments were implemented in PyTorch and run on NVIDIA L40s GPUs (48 GB). Our hierarchical decoder uses convolution channels of 256, 128, and 64. We trained for 300 epochs with DiceCELoss from MONAI \cite{cardoso_monai_2022}, selecting the model with the lowest validation loss. Data augmentation included flipping, rotation, scaling, contrast adjustment, Gaussian noise, and smoothing (probability = 0.5). Optimization used Adam with a linear warmup (0 to $1\text{e}^{-4}$) and cosine decay. Feature extraction employed Hiera-L ($d_{\text{hiera}}=256$) as the image encoder and ViT-S-14 ($d_{\text{dino}}=384$) for DINOv2.


\section{Results}
\noindent\textbf{Segmentation accuracy.}
\begin{table}[tb]
    \centering
    \caption{Quantitative evaluation on cardiac ultrasound datasets under different supervision levels (1\%, 10\%, 100\%) using Dice Score (DSC) and Hausdorff Distance (HD), averaged over the left ventricle and atrium. Units: DSC (\%), HD (px). The best method is bolded, and the second-best is underlined.}
    \begin{adjustbox}{scale=0.8}
    \begin{tabular}{lcccccc|cccccc}
    \toprule
    &  \multicolumn{6}{c|}{CAMUS \cite{leclerc_deep_2019}} & \multicolumn{6}{c}{CardiacUDA \cite{yang_graphecho_2023}}\\
    
    \cmidrule(lr){2-7} \cmidrule(lr){8-13} 
    
    & \multicolumn{2}{c}{1\%} & \multicolumn{2}{c}{10\%} & \multicolumn{2}{c|}{100\%}  & \multicolumn{2}{c}{1\%} & \multicolumn{2}{c}{10\%} & \multicolumn{2}{c}{100\%}\\

    \cmidrule(lr){2-3} \cmidrule(lr){4-5} \cmidrule(lr){6-7} \cmidrule(lr){8-9} \cmidrule(lr){10-11} \cmidrule(lr){12-13} 

    & DSC $\uparrow$ & HD $\downarrow$ & DSC $\uparrow$ & HD $\downarrow$ & DSC $\uparrow$ & HD $\downarrow$ & DSC $\uparrow$ & HD $\downarrow$ & DSC $\uparrow$ & HD $\downarrow$ & DSC $\uparrow$ & HD $\downarrow$\\\midrule
    
    UNet \cite{ronneberger_u-net_2015} &  59.49 & 51.51 & 80.14 & 26.36 & 87.70 & 15.05 & 34.65 & 41.04 & 72.83 & 26.24 & 81.69 & 20.78\\
    nnUNet \cite{isensee_nnu-net_2021} & 67.02 & 54.22 & \underline{89.36} & \underline{8.83} & \underline{91.69} & \textbf{6.35} & 35.48 & 58.94 & 49.49 & 34.12 & \textbf{90.39} & 8.32 \\\midrule
    MedNeXt \cite{greenspan_mednext_2023} & 63.30 & 48.23 & 84.73 & 26.16 & 89.49 & 9.83 & 19.05 & 158.71 & 70.97 & 32.59 & 85.54 & 8.34 \\
    SwinUNETR \cite{greenspan_swinunetr-v2_2023} & 71.77 & 48.80 & 84.99 & 23.81 & 89.36 & 12.24 & 51.54 & 90.38 & 81.44 & 24.11 & 85.81 & 9.94\\\midrule
    SAMUS \cite{lin_beyond_2024} & \underline{75.00} & \underline{26.29} & 87.43 & 16.24 & 91.11 & 8.79 & \underline{63.65} & \underline{23.91} & \underline{82.10} & \textbf{10.15} & 86.14 & \underline{7.70}\\
    MedSAM2 \cite{zhu_medical_2024} & 9.75 & 137.29 & 44.24 & 42.21 & 84.76 & 12.96 & 3.18 & 154.61 & 3.82 & 151.34 & 75.78 & 12.95\\
    Ours & \textbf{81.96} & \textbf{17.16} & \textbf{90.30} & \textbf{8.44} & \textbf{92.01} & \underline{6.75} & \textbf{65.84} & \textbf{23.57} & \textbf{82.38} & \underline{11.07} & \underline{87.44} & \textbf{6.45}\\
    
    \bottomrule
    \end{tabular}
    \end{adjustbox}
    
    \label{tab:cardiac}
\end{table}
\begin{figure}[tb]
    \centering
    \includegraphics[scale=0.138]{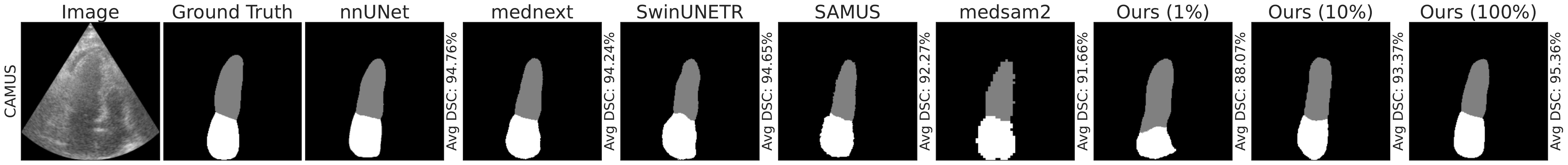} \\
    \includegraphics[scale=0.138]{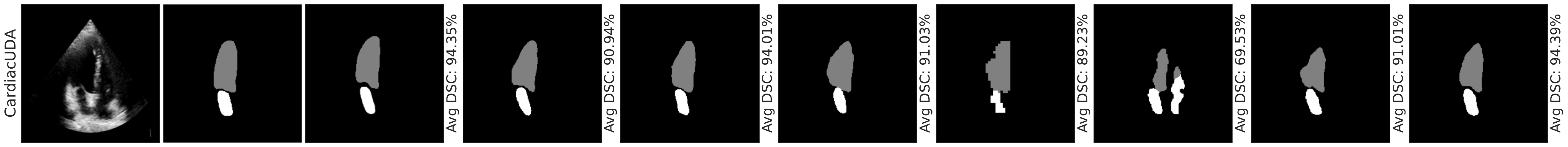}
    \caption{Qualitative evaluation on cardiac ultrasound datasets, with average DSC computed for each method against ground truth of left ventricle (gray) and atrium (white).}
    \label{fig:cardiac}
\end{figure}
\begin{table}[tb]
    \centering
    \caption{Quantitative evaluation on thyroid ultrasound datasets using Dice Score (DSC) and 95th-percentile Hausdorff Distance (HD95).
    Units: DSC (\%), HD95 (px). 
    }
    \begin{adjustbox}{scale=0.8}
    \begin{tabular}{lcc|cc|cc|cc|cc}
    \toprule
    &  \multicolumn{2}{c|}{TN3K \cite{gong_multi-task_2021}} & \multicolumn{2}{c|}{DDTI \cite{pedraza_open_2015}} & \multicolumn{2}{c|}{Stanford \cite{noauthor_thyroid_nodate}} & \multicolumn{2}{c|}{TG3K \cite{gong_multi-task_2021}} & \multicolumn{2}{c}{In-house}\\
    

    \cmidrule(lr){2-3} \cmidrule(lr){4-5} \cmidrule(lr){6-7} \cmidrule(lr){8-9} \cmidrule(lr){10-11}

    & DSC $\uparrow$ & HD95 $\downarrow$ & DSC $\uparrow$ & HD95 $\downarrow$ & DSC $\uparrow$ & HD95 $\downarrow$ & DSC $\uparrow$ & HD95 $\downarrow$ & DSC $\uparrow$ & HD95 $\downarrow$\\\midrule
    
    UNet \cite{ronneberger_u-net_2015} & 67.93 & 41.26 & 48.43 & 52.60 & 89.09 & 18.96 & 70.27 & 65.88 & 68.50 & 60.72\\
    nnUNet \cite{isensee_nnu-net_2021} & \underline{85.13} & \underline{17.24} & 73.26 & 40.43 & 96.92 & 2.72 & \underline{77.50} & 22.52 & 79.38 & \underline{18.59} \\\midrule
    MedNeXt \cite{greenspan_mednext_2023} & 70.77 & 32.58 & 71.59 & 40.02& \underline{97.33} & 2.59 & 77.13 & 40.42 & \underline{79.75} & 20.11\\
    SwinUNETR \cite{greenspan_swinunetr-v2_2023} & 71.84 & 37.89 & 70.59 & 41.76 & \textbf{97.63} & \underline{2.52} & 69.72 & 43.37 & 76.21 & 39.52\\\midrule
    SAMUS \cite{lin_beyond_2024} & 82.60 & 18.18 & \underline{77.48} & \underline{33.53} & 96.36 & 2.70 & 33.18 & 58.15 & 78.31 & 32.61 \\
    MedSAM2 \cite{zhu_medical_2024} & 69.02 & 31.43 & 69.69 & 39.02 & 87.20 & 9.58 & 75.86 & \underline{21.10} & 79.47 & 18.70 \\
    Ours & \textbf{86.01} & \textbf{15.43} & \textbf{81.52} & \textbf{26.68} & \underline{97.33} & \textbf{2.23} & \textbf{83.04} & \textbf{12.55} & \textbf{82.59} & \textbf{17.11}\\
    \bottomrule
    \end{tabular}
    \end{adjustbox}
    \label{tab:thyroid}
\end{table}
\begin{figure}[tb]
    \centering
    \begin{minipage}{0.60\textwidth}
        \centering
        \includegraphics[scale=0.103]{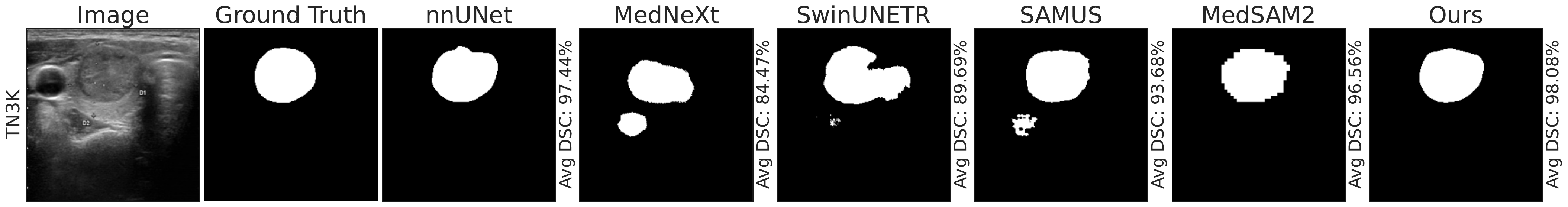} \\
        \includegraphics[scale=0.103]{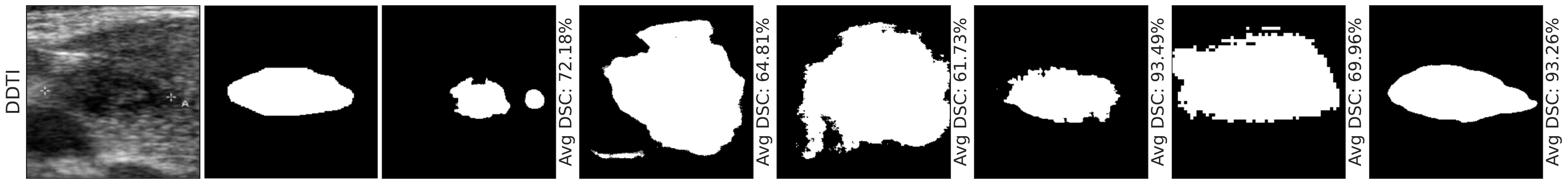} \\
        \includegraphics[scale=0.103]{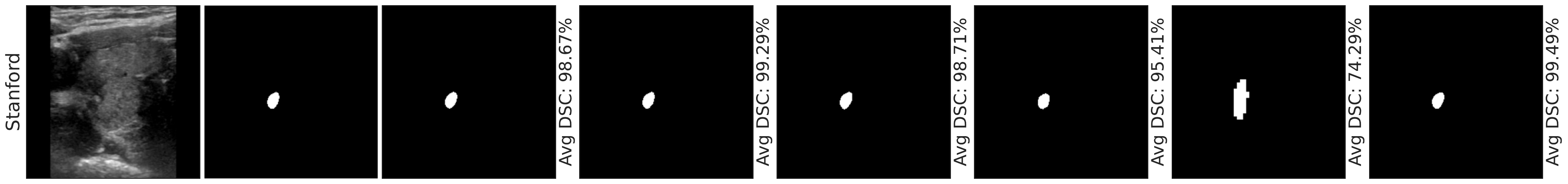} \\
        \includegraphics[scale=0.103]{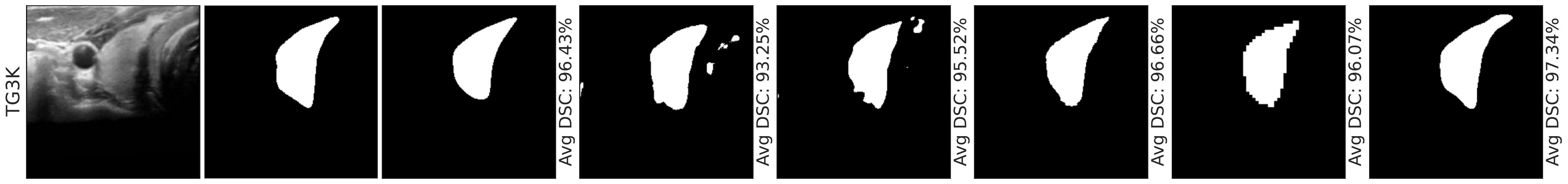} \\
        \caption{Qualitative evaluation on thyroid ultrasound datasets, with DSC computed for each method against ground truth. First three rows correspond to nodules, while the last row represents gland.}
        \label{fig:thyroid}

        \vspace{8pt}
        \includegraphics[scale=0.12]{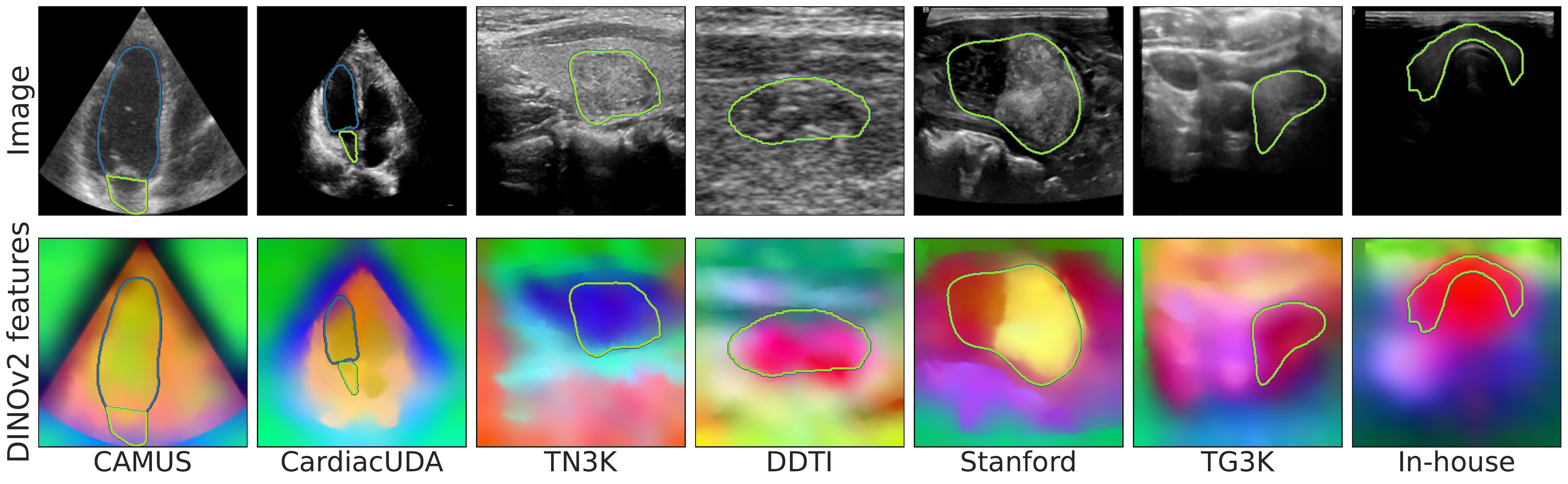}
        \caption{Qualitative evaluation of the semantic expressiveness of DINOv2 features visualized by plotting the first three principal components. Ground truth labels are overlaid as contours for reference.}
        \label{fig:dinov2}
    \end{minipage}
    \hfill
    \begin{minipage}{0.38\textwidth}
        \centering
        \includegraphics[scale=0.12]{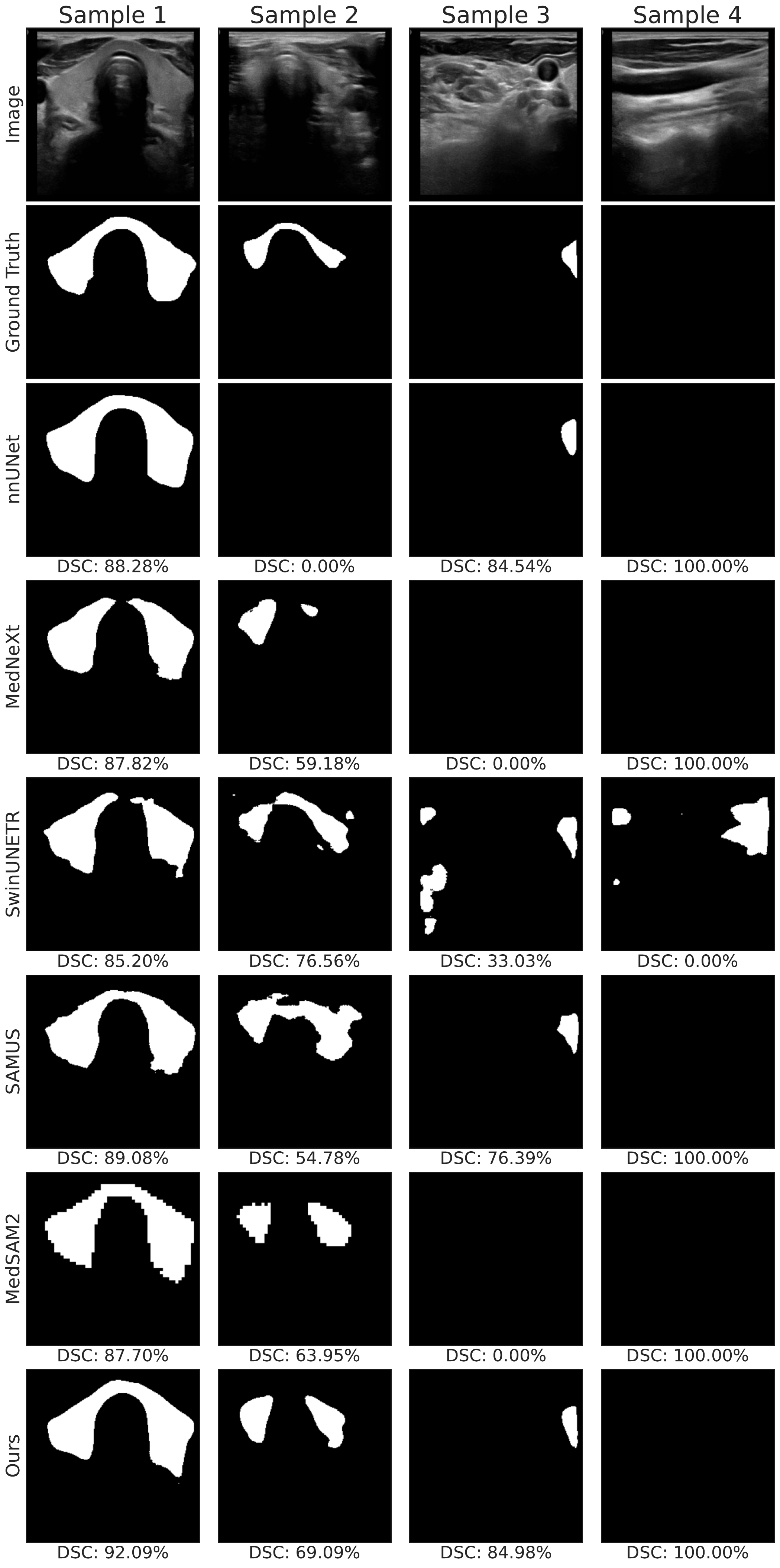}
        \caption{Qualitative evaluation on our in-house datasets, with DSC computed against ground truth.}
        \label{fig:uih}
    \end{minipage}
\end{figure}
We present our main quantitative analysis of segmentation performance in \cref{tab:cardiac} for cardiac ultrasound and \cref{tab:thyroid} for thyroid ultrasound. In both tables, our proposed approach (last row) consistently achieves the best overall performance, ranking first in 9 out of 11 metrics across DSC and HD/HD95, demonstrating the effectiveness and robustness.
We further extend our comparison with SOTA methods on CAMUS (\cref{tab:camus_sota}) and TN3K (\cref{tab:tn3k_sota}). Our proposed approach consistently outperforms existing SOTA methods in region overlap metrics in terms of ACC, DSC, and IoU while maintaining competitive performance in distance-based metrics in terms of HD95 and ASD across both datasets. These results further validate the effectiveness of our approach.

\noindent\textbf{Data efficiency and adaptability.}
Furthermore, as shown in \cref{tab:cardiac}, our approach remains highly effective even under limited supervision, significantly outperforming established baselines when trained with only 1\% and 10\% of the training data. Specifically, our method surpasses state-of-the-art nnUNet by an average of $\sim$23.6\% in DSC and outperforms transformer-based approaches like SwinUNETR by $\sim$7.7\%, while also demonstrating superiority over SAM-based approaches such as SAMUS. This validates the adaptability of our method, which achieves strong performance without requiring extensive labeled data. The qualitative results in \cref{fig:cardiac} further support this, showing smoother boundaries and better alignment with the ground truth compared to other approaches. 

\noindent\textbf{Cross-dataset generalization.}
Our method demonstrates strong generalization capability when trained on TN3K and tested on DDTI, as shown in the second column of \cref{tab:thyroid}, surpassing the second-best method (SAMUS) by more than 4\%. The qualitative results in \cref{fig:thyroid} further validate this improvement. 
%


\noindent\textbf{In-house evaluation.}
\begin{table*}[tb]
    \centering
    \begin{minipage}{0.48\textwidth}
        \centering
        \caption{Comparison of left ventricle segmentation with SOTA methods on CAMUS dataset. Units: DSC (\%), IoU (\%), HD95 (mm), ASD (mm).
        }
        \vspace{-5pt}
        \begin{adjustbox}{scale=0.8}
            \begin{tabular}{lcccc}
                \toprule
                Method & DSC $\uparrow$ & IoU $\uparrow$ & HD95 $\downarrow$ & ASD $\downarrow$ \\
                \midrule
                SwinUNet & 88.84 & 80.33 & 6.10 & 2.60 \\
                H2Former & 91.31 & 84.30 & 5.27 & 2.05 \\
                \midrule
                MedSAM & 85.42 & 75.14 & 8.42 & 3.34 \\
                MSA & 88.03 & 78.98 & 7.53 & 2.85 \\
                SAMed & 87.45 & 78.14 & 9.17 & 3.10 \\
                SonoSAM & 89.80 & 81.79 & 6.60 & 2.45 \\
                MemSAM & \underline{93.31} & \underline{87.61} & \textbf{3.82} & \textbf{1.57}\\ 
                Ours & \textbf{93.80} & \textbf{88.49} & \underline{4.80} & \underline{1.90}\\
                \bottomrule
            \end{tabular}
        \end{adjustbox}
        \label{tab:camus_sota}
    \end{minipage}
    \hfill
    \begin{minipage}{0.49\textwidth}
        \centering
        \vspace{-18pt}
        \caption{Comparison with SOTA methods on the TN3K dataset. Accuracy is denoted as ACC. Units: ACC (\%), DSC (\%), IoU (\%), HD95 (px).}
        \begin{adjustbox}{scale=0.8}
            \begin{tabular}{lcccc}
                \toprule
                Method & ACC $\uparrow$ & IoU $\uparrow$ & DSC$\uparrow$ & HD95$\downarrow$  \\
                \midrule
                TRFE & 96.71 & 68.33 & 81.91 & 17.96 \\
                SegNet & 96.72 & 66.54 & 79.91 & 17.13 \\
                DeepLabv3 & 97.19 & 70.60 & 82.77 & 13.92 \\
                TRFE+ & 97.04 & 71.38 & 83.30 & \underline{13.23} \\
                SHAN & \underline{96.73} & \underline{73.59} & \underline{84.61} & \textbf{4.05} \\
                Ours & \textbf{97.60} & \textbf{78.13} & \textbf{86.01} & 15.42\\
                \bottomrule
            \end{tabular}
        \end{adjustbox}
        \label{tab:tn3k_sota}
    \end{minipage}
    \hfill
    \begin{minipage}{0.48\textwidth}
        \centering
        \vspace{5pt}
        \caption{Quantitative image-level evaluation on in-house ultrasound datasets. All values are reported as percentages (\%).}
        \label{tab:uih_detection}
        \begin{adjustbox}{scale=0.65}
            \begin{tabular}{lcccc}
                \toprule
                Model & Precision $\uparrow$ & Recall $\uparrow$ & Specificity $\uparrow$ & F1-Score $\uparrow$\\
                \midrule
                UNet & 64.75 & 93.92 & 12.41 & 96.86 \\
                nnUNet & 91.97 & 93.48 & 86.02 & 96.63 \\ \midrule
                MedNeXt & 93.90 & 86.96 & 90.32 & 93.03 \\
                SwinUNETR & 62.09 & 95.61 & 0.00 & 97.75\\ \midrule
                SAMUS & \textbf{99.71} & 98.26 & \textbf{99.50} & 99.12 \\
                MedSAM2 & 96.67 & 85.61 & 94.95 & 92.25 \\
                Ours & 87.21 & \textbf{98.41} & 75.27 & \textbf{99.20} \\
                \bottomrule
            \end{tabular}
        \end{adjustbox}
    \end{minipage}
    \hfill
    \begin{minipage}{0.49\textwidth}
        \centering
        \vspace{-5pt}
        \caption{Ablation study on CAMUS: (1) Hierarchical decoder (H-Dec.), (2) Hiera adapter (H-Adp.), (3) DINOv2 feature integration, and (4) Feature interleaving. Units: DSC (\%), HD (px)}
        \begin{adjustbox}{scale=0.73}
            \begin{tabular}{cccc|cc}
                \toprule
                H-Dec. & H-Adp. & DINOv2 & Interleave & DSC $\uparrow$ & HD $\downarrow$ \\
                \midrule
                \xmark & \xmark & \xmark & \xmark & 78.23 & 30.98\\
                \cmark & \xmark & \xmark & \xmark & 90.85 & 9.42\\
                \cmark & \cmark & \xmark & \xmark & 91.89 & 6.84\\
                \cmark & \cmark & \cmark & \xmark & 91.90 & 6.78\\
                \cmark & \cmark & \cmark & \cmark & \textbf{92.01} & \textbf{6.75}\\
                \bottomrule
            \end{tabular}
        \end{adjustbox}
        \label{tab:ablation}
    \end{minipage}
\end{table*}

To assess real-world applicability, we computed image-level detection metrics, including precision, recall, specificity, and F1-score, on our in-house dataset (\cref{tab:uih_detection}). This dataset presents a significant challenge since the thyroid gland may not always be present in the image, as in \cref{fig:uih} (e.g., sample 4). We define the classification metrics as follows: (1) TP: Prediction exists when the ground truth (GT) exists, and IoU $>$ threshold. (2) FN: GT exists, but the prediction is missing or IoU $\leq$ threshold. (3) TN: Both GT and prediction are empty. (4) FP: GT is empty, but the prediction exists. We set threshold=0.25. Our method achieves the highest recall (98.41\%) and F1-score (99.20\%), demonstrating its ability to detect nearly all segmentations while maintaining a strong balance with precision (87.21\%) and specificity (75.27\%). Compared to others, nnUNet and MedNeXt exhibit strong specificity but slightly lower recall, while SwinUNETR and UNet tend to over-segment, leading to poor specificity. Notably, SAMUS achieves the highest precision (99.71\%) and specificity (99.50\%), while our method remains comparable, with superior recall and F1-score, indicating its robustness in detecting segmentations with minimal missing regions.
\noindent\textbf{Inference speed.}  
Our approach achieves $\sim$30 FPS on a single GPU and $\sim$77 FPS with TensorRT when tested on $224 \times 224$ images.

\noindent\textbf{Ablation study.}
To assess the impact of each component, we perform an ablation study in \cref{tab:ablation}. The first row serves as a baseline with a convolutional decoder processing the finest-scale SAM2 features without adaptation. The second row adds our hierarchical decoder for multi-scale decoding. The third row incorporates the Hiera adapter for enhanced feature refinement, while the fourth row integrates DINOv2 features via concatenation. Finally, our full model (last row) with feature interleaving achieves the best DSC and HD, demonstrating the effectiveness of our design choices. The integration of DINOv2 is further validated in \cref{fig:dinov2}, where the first three principal components align with ground truth anatomical structures, highlighting its role in capturing meaningful semantics for segmenting indistinct ultrasound boundaries.

\section{Conclusion}
\label{sec:conclusion}
We propose a method for adapting vision foundation models to ultrasound segmentation, introducing a Hiera adapter for hierarchical feature extraction and integrating DINOv2 for enhanced visual representation. The enriched multi-scale features are then decoded to generate precise segmentation masks. Evaluated on seven ultrasound datasets, our approach consistently outperforms baselines and achieves SOTA performance. Notably, our model enables real-time segmentation at $\sim$77 FPS with TensorRT. This work demonstrates the potential of combining foundation models with supervised training on limited data for high-quality real-time segmentation, paving the way for broader applications in medical imaging. Future work will explore extensions to video and 3D imaging. 

%
%
%
%
%
%
\bibliographystyle{splncs04}
\bibliography{references_short}

\begin{thebibliography}{10}
\providecommand{\url}[1]{\texttt{#1}}
\providecommand{\urlprefix}{URL }
\providecommand{\doi}[1]{https://doi.org/#1}

\bibitem{noauthor_thyroid_nodate}
Thyroid {Ultrasound} {Cine}-clip {\textbar} {Center} for {AI} in {Medicine} \& {Imaging}

\bibitem{amiri_fine-tuning_2020}
Amiri, M., et~al.: Fine-{Tuning} {U}-{Net} for {Ultrasound} {Image} {Segmentation}: {Different} {Layers}, {Different} {Outcomes}. IEEE Trans. Ultrason. Ferroelectr. Freq. Control

\bibitem{ansari_advancements_2024}
Ansari, M.Y., et~al.: Advancements in {Deep} {Learning} for {B}-{Mode} {Ultrasound} {Segmentation}: {A} {Comprehensive} {Review}. IEEE Trans. Emerg. Top. Comput. Intell.

\bibitem{ansari_dense-psp-unet_2023}
Ansari, M.Y., et~al.: Dense-{PSP}-{UNet}: {A} neural network for fast inference liver ultrasound segmentation. Comput. Biol. Med.  \textbf{153},  106478 (Feb 2023)

\bibitem{awais_foundation_2025}
Awais, M., et~al.: Foundation {Models} {Defining} a {New} {Era} in {Vision}: a {Survey} and {Outlook}. IEEE Trans. Pattern Anal. Mach. Intell. pp. 1--20 (2025)

\bibitem{bernard_deep_2018}
Bernard, O., et~al.: Deep {Learning} {Techniques} for {Automatic} {MRI} {Cardiac} {Multi}-{Structures} {Segmentation} and {Diagnosis}: {Is} the {Problem} {Solved}? IEEE Trans. Med. Imaging  \textbf{37}(11),  2514--2525 (Nov 2018)

\bibitem{cardoso_monai_2022}
Cardoso, M.J., et~al.: {MONAI}: {An} open-source framework for deep learning in healthcare (Nov 2022), arXiv:2211.02701 [cs]

\bibitem{chen_transunet_2021}
Chen, J., et~al.: {TransUNet}: {Transformers} {Make} {Strong} {Encoders} for {Medical} {Image} {Segmentation} (Feb 2021), arXiv:2102.04306 [cs]

\bibitem{deng_memsam_2024}
Deng, X., et~al.: {MemSAM}: {Taming} {Segment} {Anything} {Model} for {Echocardiography} {Video} {Segmentation}. In: {CVPR}. pp. 9622--9631. IEEE (Jun 2024)

\bibitem{dorent_patient-specific_2024}
Dorent, R., et~al.: Patient-{Specific} {Real}-{Time} {Segmentation} in {Trackerless} {Brain} {Ultrasound}. In: {MICCAI} 2024. pp. 477--487. Springer Nature Switzerland (2024)

\bibitem{dosovitskiy_image_2020}
Dosovitskiy, A., et~al.: An {Image} is {Worth} 16x16 {Words}: {Transformers} for {Image} {Recognition} at {Scale} (Oct 2020)

\bibitem{gong_multi-task_2021}
Gong, H., et~al.: Multi-{Task} {Learning} {For} {Thyroid} {Nodule} {Segmentation} {With} {Thyroid} {Region} {Prior}. In: {ISBI}. pp. 257--261 (Apr 2021)

\bibitem{greenspan_swinunetr-v2_2023}
He, Y., et~al.: {SwinUNETR}-{V2}: {Stronger} {Swin} {Transformers} with {Stagewise} {Convolutions} for {3D} {Medical} {Image} {Segmentation}. In: {MICCAI}. Springer (2023)

\bibitem{howell_deep_2024}
Howell, L., et~al.: Deep learning for real-time multi-class segmentation of artefacts in lung ultrasound. Ultrasonics  \textbf{140},  107251 (May 2024)

\bibitem{hu_lora_2021}
Hu, E.J., et~al.: {LoRA}: {Low}-{Rank} {Adaptation} of {Large} {Language} {Models}

\bibitem{isensee_nnu-net_2021}
Isensee, F., et~al.: {nnU}-{Net}: A self-configuring method for deep learning-based biomedical image segmentation. Nat. Methods  \textbf{18}(2),  203--211 (Feb 2021)

\bibitem{avidan_visual_2022}
Jia, M., et~al.: Visual {Prompt} {Tuning}. In: Comput. Vis. – {ECCV} 2022, vol. 13693, pp. 709--727. Springer Nature Switzerland, Cham (2022)

\bibitem{jiang_robotic_2023}
Jiang, Z., et~al.: Robotic ultrasound imaging: {State}-of-the-art and future perspectives. Med. Image Anal.  \textbf{89},  102878 (Oct 2023)

\bibitem{kirillov_segment_2023}
Kirillov, A., et~al.: Segment {Anything}. In: {ICCV}. pp. 3992--4003. IEEE (Oct 2023)

\bibitem{leclerc_deep_2019}
Leclerc, S., et~al.: Deep {Learning} for {Segmentation} {Using} an {Open} {Large}-{Scale} {Dataset} in {2D} {Echocardiography}. IEEE Trans. Med. Imaging  \textbf{38},  2198--2210

\bibitem{lester_power_2021}
Lester, B., et~al.: The {Power} of {Scale} for {Parameter}-{Efficient} {Prompt} {Tuning}. In: {EMNLP}. pp. 3045--3059. {ACL}

\bibitem{li_mvitv2_2022}
Li, Y., et~al.: {MViTv2}: {Improved} {Multiscale} {Vision} {Transformers} for {Classification} and {Detection}. In: ({CVPR}). pp. 4794--4804. IEEE (Jun 2022)

\bibitem{lin_beyond_2024}
Lin, X., et~al.: Beyond {Adapting} {SAM}: {Towards} {End}-to-{End} {Ultrasound} {Image} {Segmentation} via {Auto} {Prompting} (Jul 2024), arXiv:2309.06824 [cs]

\bibitem{loizou_review_2014}
Loizou, C.P.: A review of ultrasound common carotid artery image and video segmentation techniques. Med. Biol. Eng. Comput.  \textbf{52}(12),  1073--1093 (Dec 2014)

\bibitem{looney_fully_2018}
Looney, P., et~al.: Fully automated, real-time {3D} ultrasound segmentation to estimate first trimester placental volume using deep learning. JCI Insight  \textbf{3}(11)

\bibitem{ma_segment_2024}
Ma, J., et~al.: Segment {Anything} in {Medical} {Images} and {Videos}: {Benchmark} and {Deployment} (Aug 2024), arXiv:2408.03322 [eess]

\bibitem{michailovich_despeckling_2006}
Michailovich, O.V., et~al.: Despeckling of {Medical} {Ultrasound} {Images}. IEEE Trans. Ultrason. Ferroelectr. Freq. Control  \textbf{53}(1),  64--78 (Jan 2006)

\bibitem{oquab_dinov2_2023}
Oquab, M., et~al.: {DINOv2}: {Learning} {Robust} {Visual} {Features} without {Supervision}. Trans. Mach. Learn. Res.  (Jul 2023)

\bibitem{ou_rtseg-net_2024}
Ou, Z., et~al.: {RTSeg}-net: {A} lightweight network for real-time segmentation of fetal head and pubic symphysis from intrapartum ultrasound images. Comput. Biol. Med.  \textbf{175},  108501 (Jun 2024)

\bibitem{pedraza_open_2015}
Pedraza, L., et~al.: An open access thyroid ultrasound image database. In: 10th Int. Symp. Med. Inf. Process. Anal. vol.~9287, pp. 188--193. SPIE (Jan 2015)

\bibitem{qiu_controlling_2023}
Qiu, Z., et~al.: Controlling {Text}-to-{Image} {Diffusion} by {Orthogonal} {Finetuning}

\bibitem{ravi_sam_2024}
Ravi, N., et~al.: {SAM} 2: {Segment} {Anything} in {Images} and {Videos} (Oct 2024)

\bibitem{ravishankar_sonosam_2023}
Ravishankar, H., et~al.: {SonoSAM} - {Segment} {Anything} on {Ultrasound} {Images}. In: Simplifying {Medical} {Ultrasound}. pp. 23--33. Springer Nature Switzerland

\bibitem{rebuffi_learning_2017}
Rebuffi, S.A., et~al.: Learning multiple visual domains with residual adapters. In: Adv. Neural Inf. Process. Syst. ({NeurIPS}). vol.~30. Curran Associates, Inc. (2017)

\bibitem{ronneberger_u-net_2015}
Ronneberger, O., et~al.: U-{Net}: {Convolutional} {Networks} for {Biomedical} {Image} {Segmentation}. In: {MICCAI} 2015. pp. 234--241. Springer (2015)

\bibitem{greenspan_mednext_2023}
Roy, S., et~al.: {MedNeXt}: {Transformer}-{Driven} {Scaling} of {ConvNets} for {Medical} {Image} {Segmentation}. In: {MICCAI} 2023, vol. 14223, pp. 405--415. Springer

\bibitem{ryali_hiera_2023}
Ryali, C., et~al.: Hiera: {A} {Hierarchical} {Vision} {Transformer} without the {Bells}-and-{Whistles}. In: {ICML}. PMLR (Jul 2023)

\bibitem{salcudean_robot-assisted_2022}
Salcudean, S.E., et~al.: Robot-{Assisted} {Medical} {Imaging}: {A} {Review}. Proc. IEEE

\bibitem{steitz_adapters_2024}
Steitz, J.M.O., et~al.: Adapters {Strike} {Back}. In: {CVPR}. IEEE (Jun 2024)

\bibitem{stock_segment_2024}
Stock, R., et~al.: Segment {Anything} in {Medical} {Images} with {nnUNet} (Jun 2024)

\bibitem{ta_multi-task_2024}
Ta, K., et~al.: Multi-{Task} {Learning} for {Motion} {Analysis} and {Segmentation} in {3D} {Echocardiography}. IEEE Trans. Med. Imaging  \textbf{43}(5),  2010--2020 (May 2024)

\bibitem{tong_cambrian-1_2024}
Tong, S., et~al.: Cambrian-1: {A} {Fully} {Open}, {Vision}-{Centric} {Exploration} of {Multimodal} {LLMs} (Dec 2024), arXiv:2406.16860 [cs]

\bibitem{vaze_low-memory_2020}
Vaze, S., et~al.: Low-{Memory} {CNNs} {Enabling} {Real}-{Time} {Ultrasound} {Segmentation} {Towards} {Mobile} {Deployment}. IEEE J. Biomed. Health Inform.  \textbf{24}(4),  1059--1069

\bibitem{webb_comparing_2021}
Webb, J.M., et~al.: Comparing deep learning-based automatic segmentation of breast masses to expert interobserver variability in ultrasound imaging. Comput. Biol. Med.  \textbf{139},  104966 (Dec 2021)

\bibitem{wu_cross-image_2023}
Wu, H., et~al.: Cross-{Image} {Dependency} {Modeling} for {Breast} {Ultrasound} {Segmentation}. IEEE Trans. Med. Imaging  \textbf{42}(6),  1619--1631 (Jun 2023)

\bibitem{yang_graphecho_2023}
Yang, J., et~al.: {GraphEcho}: {Graph}-{Driven} {Unsupervised} {Domain} {Adaptation} for {Echocardiogram} {Video} {Segmentation}. In: {ICCV}. pp. 11844--11853. IEEE

\bibitem{yeh_navigating_2024}
Yeh, S.Y., et~al.: Navigating {Text}-{To}-{Image} {Customization}: {From} {LyCORIS} {Fine}-{Tuning} to {Model} {Evaluation} (Mar 2024), arXiv:2309.14859 [cs]

\bibitem{linguraru_shan_2024}
Zhang, R., et~al.: {SHAN}: {Shape} {Guided} {Network} for {Thyroid} {Nodule} {Ultrasound} {Cross}-{Domain} {Segmentation}. In: {MICCAI}, vol. 15004, pp. 732--741. Springer (2024)

\bibitem{leonardis_adaptive_2025}
Zhang, X., et~al.: Adaptive {Correspondence} {Scoring} for {Unsupervised} {Medical} {Image} {Registration}. In: {ECCV} 2024, vol. 15096, pp. 76--92. Springer

\bibitem{linguraru_heteroscedastic_2024}
Zhang, X., et~al.: Heteroscedastic {Uncertainty} {Estimation} {Framework} for {Unsupervised} {Registration}. In: {MICCAI} 2024, pp. 651--661. Springer

\bibitem{zhou_cardiac_2017}
Zhou, R., et~al.: Cardiac atlas development and validation for automatic segmentation of cardiac substructures. Radiother. Oncol.  \textbf{122}(1),  66--71 (Jan 2017)

\bibitem{zhu_medical_2024}
Zhu, J., et~al.: Medical {SAM} 2: {Segment} medical images as video via {Segment} {Anything} {Model} 2 (Dec 2024), arXiv:2408.00874 [cs]

\end{thebibliography}

\end{document}